\documentclass{article}
\usepackage{arxiv}
\usepackage[T1]{fontenc}
\usepackage{microtype}
\usepackage{amssymb,amsmath,amsthm}
\usepackage{graphicx}
\usepackage{booktabs}
\usepackage{array}
\usepackage{algorithm}
\usepackage{algpseudocode}
\usepackage{float}
\usepackage{enumitem}
\usepackage{url}
\usepackage[round]{natbib}
\usepackage{xcolor}
\usepackage[colorlinks=true,linkcolor=blue!60!black,citecolor=blue!60!black,urlcolor=blue!60!black]{hyperref}

\hypersetup{pdftitle={Understanding Unified VLMs: From Attention Probing to Guidance Tuning for Multimodal Generation},
  pdfauthor={Zhengyang Luo, Xiaoyu Qiu, Zhiheng Tan, Yuchen Liang, Yingbin Liang, Liyue Shen}}
\newtheorem{theorem}{Theorem}


\title{Understanding Unified VLMs: From Attention Probing to Guidance Tuning for Multimodal Generation}

\author{%
  Zhengyang Luo$^{1}$ \quad Xiaoyu Qiu$^{1}$ \quad Zhiheng Tan$^{2}$ \quad Yuchen Liang$^{3}$ \quad Yingbin Liang$^{2}$ \quad Liyue Shen$^{1}$ \\[4pt]
  {\normalfont $^{1}$University of Michigan, Ann Arbor \quad $^{2}$The Ohio State University \quad $^{3}$Indiana University Indianapolis}
}
\renewcommand{\shorttitle}{Understanding Unified VLMs: From Attention Probing to Guidance Tuning}
\date{}

\begin{document}

\maketitle

\begin{abstract}
Unified vision-language models (VLMs) perform both text-to-image (T2I) and image-to-text (I2T) generation with one shared Transformer
backbone, yet how the backbone processes the condition in either direction remains poorly understood.
We study this question from the perspective of conditional guidance: we probe every attention head to show the activation difference between two branches of classifier-free guidance (CFG) with different conditional guidance.
On two unified VLMs of autoregressive and masked-diffusion paradigms, the probing reveals a subset of heads that are more sensitive to the change of condition while others are agnostic to it.
This characteristic pattern holds in both T2I and I2T generation and is largely stable across data samples, demonstrating the internal mechanism of unified VLMs.
Motivated by this, we propose a new inference-time intervention approach for unified multimodal generation,
named \textit{HeadTuned-CFG}, which uses the probed condition-change awareness to select and tune the conditional guidance at the attention-head level.
Experiments show consistent benefits in both multimodal generation directions of unified VLMs without additional training, and in
particular substantial improvements over strong baselines in T2I generation.
These results further support the key finding of our probing analysis: the observed pattern of functional specialization is an intrinsic property of unified
VLMs.
\end{abstract}

\section{Introduction}
\label{sec:intro}
Unified vision-language models (VLMs) are advancing rapidly, aiming to perform multimodal understanding and generation within a single model.
Autoregressive models such as Janus~\citep{wu2024janusdecouplingvisualencoding} and Emu3~\citep{wang2024emu3nexttokenpredictionneed}, as well as
discrete diffusion models such as Muddit~\citep{shi2026mudditliberatinggenerationtexttoimage} and MMaDA~\citep{yang2025mmadamultimodallargediffusion},
have shown strong performance on unified multimodal tasks such as text-to-image (T2I) and image-to-text (I2T) generation, narrowing the gap with task-specific models.
Despite substantial differences in their training objectives and decoding strategies, these models commonly rely on a shared
Transformer backbone to process information across modalities.

\begin{figure}[!t]
\centering
\includegraphics[width=0.98\linewidth]{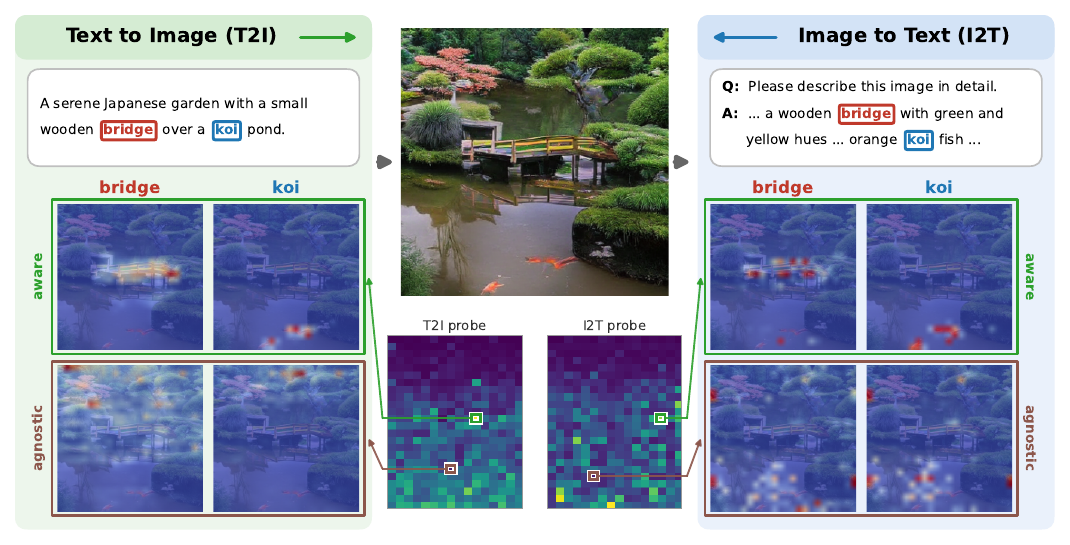}
\caption{Condition-change aware (green) and agnostic (brown) attention heads of Janus.
\textbf{Left: T2I generation}; each map shows where the generated image attends to a boxed keyword of the prompt.
\textbf{Right: I2T generation} with the same image as input; each map shows where a boxed keyword
of the generated caption attends in the input image.
\textbf{Middle: the generated (input) image and head-level condition-change awareness map} of both T2I and I2T probing
(layer by head, brighter means higher awareness), with the visualized heads marked. The Muddit counterpart is in Appendix~\ref{app:teaser-muddit}.}
\label{fig:teaser}
\end{figure}

This raises the question of how a single shared backbone supports cross-modal understanding and generation under corresponding prompt
guidance.
Research has shown that attention heads in LLMs take on identifiable roles, such as induction and retrieval
heads~\citep{olsson2022incontextlearninginductionheads, wu2024retrievalheadmechanisticallyexplains, wang2022interpretabilitywildcircuitindirect}.
Studies of VLMs have followed this line to study the understanding direction, examining how visual information is processed and
attended to inside the language model~\citep{neo2025interpretingvisualinformationprocessing, jiang2025devilsmiddlelayerslarge,
kim2025interpretingattentionheadsimagetotext}. However, the interpretability of such mechanisms on unified VLMs still remains under-explored.

In this work, we study this question through the lens of \textit{conditional guidance} at the attention-head level.
Specifically, we compare the activations of individual attention heads between the positive and negative branches of classifier-free
guidance (CFG)~\citep{ho2022classifierfreediffusionguidance}, which differ only in the conditional guidance.
The comparison reveals that a subset of heads responds strongly to the change of condition, while some barely respond, as shown in Figure~\ref{fig:teaser}.
We further find that this condition-change awareness is largely stable across samples, and that the variation is present in both T2I
and I2T generation of the two studied VLMs, revealing an intrinsic functional specialization of unified VLMs.

Motivated by these findings, we introduce a head-level guidance tuning method for multimodal generation in unified VLMs
(\textbf{HeadTuned-CFG}), which explicitly accounts for head-level differences in condition-change awareness when applying CFG.
We evaluate HeadTuned-CFG on both multimodal generation directions.
For T2I generation, HeadTuned-CFG consistently outperforms vanilla CFG and related guidance methods on standard metrics across two studied VLMs.
For I2T generation, it composes directly with CFG-style contrastive decoding such as VCD~\citep{leng2023mitigatingobjecthallucinationslarge}
and shows the potential to further improve the robustness of VLMs against hallucination.
These results further support our probing analysis as an intrinsic property, facilitating simple yet effective guidance tuning mechanism for unified multimodal generation.

In summary, our contributions are as follows:
\begin{itemize}[leftmargin=*]
    \item \textbf{Understanding via Attention Probing:} We investigate unified VLMs from the perspective of conditional guidance by
    developing a condition-change probing framework for analyzing their internal mechanism, uncovering a consistent functional
    specialization pattern of attention heads in both T2I and I2T generation.

    \item \textbf{Guidance Tuning for Multimodal Generation:}
    Motivated by the observed probing pattern, we propose HeadTuned-CFG, which selects heads based on condition-change awareness and adjusts
    the guidance strength of the selected heads through a linear scaling of their activations. We also provide a theoretical interpretation of the tuning rule under a local Gaussian model, showing how a KL projection with a prescribed positive-awareness constraint yields its interpolation/extrapolation form.

    \item \textbf{Experiments across Benchmarks and Backbones:} We demonstrate that HeadTuned-CFG consistently benefits both multimodal
    generation directions of unified VLMs, and in particular substantially outperforms vanilla CFG and related guidance methods in T2I
    generation.
\end{itemize}

\section{Preliminaries}

\subsection{Unified Vision-Language Models}

Unified VLMs typically encode multimodal inputs into discrete tokens of a shared vocabulary using a visual
tokenizer~\citep{esser2021tamingtransformershighresolutionimage} and a text tokenizer, and can be broadly divided by their generation paradigm into
autoregressive and masked-diffusion approaches.

\textbf{Autoregressive VLMs.} These models~\citep{chameleonteam2025chameleonmixedmodalearlyfusionfoundation, wang2024emu3nexttokenpredictionneed,
wu2024janusdecouplingvisualencoding} unify generation tasks under the next-token prediction paradigm. A single decoder-only Transformer is trained on interleaved image-text sequences
via causal language modeling with the standard autoregressive negative log-likelihood objective:
\begin{equation}
    \mathcal{L} = -\sum_i \log p_\theta(x_i \mid x_{<i}),
\end{equation}
where $x = (x_1, \dots, x_N)$ is a sequence of discrete image and text tokens and $x_{<i}$ denotes the tokens before position $i$.
At inference time, generation amounts to autoregressively sampling tokens from $p_\theta(x_i \mid x_{<i})$.

\textbf{Masked-Diffusion VLMs.} These models~\citep{shi2026mudditliberatinggenerationtexttoimage, yang2025mmadamultimodallargediffusion} instead
formulate multimodal generation as a discrete denoising process.
Let $x_0$ be the clean token sequence. The noisy sequence $x_t$ at step $t \in [0, 1]$ is obtained by replacing each token of $x_0$ independently
with a special \texttt{[MASK]} token with probability $\beta(t)$, where the noise schedule $\beta(t)$ increases from $\beta(0)=0$ to
$\beta(1)=1$. The Transformer is trained to recover the original tokens from their corrupted counterparts via a masked denoising objective:
\begin{equation}
    \mathcal{L} = -\sum_t \mathbb{E}_{x_t} \!\left[ \log p_\theta(x_0 \mid x_t) \right].
\end{equation}
At inference time, generation proceeds by iteratively unmasking tokens starting from a fully masked initial sequence.

\subsection{Classifier-Free Guidance}

Classifier-free guidance (CFG)~\citep{ho2022classifierfreediffusionguidance} enhances conditional generation by running two forward passes and
extrapolating between their outputs:
a \emph{positive} branch conditioned on the condition and a \emph{negative} branch conditioned on a corrupted condition.
Denoting their outputs by $\hat{\epsilon}_{\mathrm{pos}}$ and $\hat{\epsilon}_{\mathrm{neg}}$, the guided output is
\begin{equation}
    \hat{\epsilon}_{\mathrm{CFG}} = \hat{\epsilon}_{\mathrm{neg}} + w\,\bigl(\hat{\epsilon}_{\mathrm{pos}} - \hat{\epsilon}_{\mathrm{neg}}\bigr),
\label{eq:cfg}
\end{equation}
where $w$ is the guidance scale.
With $w>1$, the output is pushed along the direction in which the two branches differ, which promotes the content specific to the condition
and suppresses the content that would also be generated without it.

\subsection{Multi-Head Attention}
\label{sec:mha}

The multi-head attention mechanism~\citep{vaswani2023attentionneed} splits the query, key, and value matrices into $H_l$ parallel heads, each operating in its own learned projection space. At the $l$-th layer of the
model, the input projections for head $h \in \{1, \dots, H_l\}$ take the form
\begin{equation}
    \mathbf{Q}_{l,h} = \mathbf{Q}_l \mathbf{W}_{l,h}^Q, \qquad
    \mathbf{K}_{l,h} = \mathbf{K}_l \mathbf{W}_{l,h}^K, \qquad
    \mathbf{V}_{l,h} = \mathbf{V}_l \mathbf{W}_{l,h}^V,
\end{equation}
with $\mathbf{Q}_l, \mathbf{K}_l, \mathbf{V}_l \in \mathbb{R}^{N \times d}$ denoting the shared query, key, and value activations, and $\mathbf{W}_{l,h}^Q, \mathbf{W}_{l,h}^K,
\mathbf{W}_{l,h}^V \in \mathbb{R}^{d \times \bar{d}}$ denoting the per-head learnable projections of width $\bar{d}$.

Each head then independently produces an attention pattern and aggregates the values:
\begin{equation}
    \mathbf{A}_{l,h} = \mathrm{Softmax}\!\left(\frac{\mathbf{Q}_{l,h} \mathbf{K}_{l,h}^\top}{\sqrt{\bar{d}}}\right), \qquad
    \mathbf{O}_{l,h} = \mathbf{A}_{l,h} \mathbf{V}_{l,h}.
\end{equation}

The outputs of all heads $\{\mathbf{O}_{l,h}\}_{h=1}^{H_l}$ are concatenated and mapped back into the model dimension via an output projection matrix $\mathbf{W}_l^O \in \mathbb{R}^{(H_l \bar{d}) \times d}$:
\begin{equation}
    \mathrm{MultiHead}(\mathbf{Q}_l, \mathbf{K}_l, \mathbf{V}_l) = \mathrm{Concat}\!\left(\mathbf{O}_{l,1}, \dots, \mathbf{O}_{l,H_l}\right) \mathbf{W}_l^O.
\end{equation}

\subsection{Related Work}
\label{sec:related}

\textbf{Classifier-Free Guidance.} CFG was proposed for conditional generation in continuous diffusion
models~\citep{ho2022classifierfreediffusionguidance} and is now extended to the T2I generation of unified
VLMs~\citep{wu2024janusdecouplingvisualencoding, shi2026mudditliberatinggenerationtexttoimage}. For I2T generation, CFG-style contrastive
decoding, such as Visual Contrastive Decoding (VCD)~\citep{leng2023mitigatingobjecthallucinationslarge} and its
variants~\citep{favero2024multimodalhallucinationcontrolvisual, wang2024mitigatinghallucinationslargevisionlanguage}, builds the negative
branch from a perturbed image and contrasts it with the original one to suppress object hallucination.
Several studies improve CFG by scheduling the guidance scale across generation timesteps~\citep{wang2024analysisclassifierfreeguidanceweight, kynkaanniemi2024applyingguidancelimitedinterval}.
However, these approaches treat the model as a black box and apply a global scale, rarely examining the underlying architectural mechanisms, such as the roles of individual attention heads.

Another line of work intervenes inside the architecture: Perturbed-Attention Guidance
(PAG)~\citep{ahn2025selfrectifyingdiffusionsamplingperturbedattention} and related
methods~\citep{hong2023improvingsamplequalitydiffusion, hong2024smoothedenergyguidanceguiding, ahn2025perturbdesignperturbationguidance} take a
forward branch perturbed at intermediate attention as the reference and apply the same linear extrapolation as CFG, which extends guidance to tasks without explicit
conditioning, such as inverse problems and inpainting. They do not outperform CFG on common conditional generation, and are therefore
mainly used as a substitute for CFG where it is inapplicable or stacked on CFG to improve it. Moreover, designed for the diffusion
paradigm, they cannot be equivalently adapted to autoregressive models.

\textbf{Unified-VLM Interpretability.} Attention heads in LLMs take on identifiable roles: induction heads implement in-context
copying~\citep{olsson2022incontextlearninginductionheads}, retrieval heads fetch information from long contexts~\citep{wu2024retrievalheadmechanisticallyexplains},
and a small circuit of heads implements a specific task~\citep{wang2022interpretabilitywildcircuitindirect}. Interpretability studies of
VLMs have followed this line in the understanding direction: \citet{neo2025interpretingvisualinformationprocessing} trace how visual tokens
are processed across the layers of the language model, \citet{jiang2025devilsmiddlelayerslarge} relate object hallucination to the attention
paid to image tokens in the middle layers, and \citet{kim2025interpretingattentionheadsimagetotext} attribute the image-to-text information
flow to a subset of heads. However, the interpretability of such mechanisms on unified VLMs remains poorly explored.

\section[Understanding Condition-change Awareness of Attention Heads: A Probing-based Analysis]{Understanding Condition-change Awareness of Attention Heads:\\ A Probing-based Analysis}
\label{sec:awareness}
To understand how individual attention heads behave differently under CFG, we probe the response of attention heads to the change of the
condition on two unified VLMs: Muddit~\citep{shi2026mudditliberatinggenerationtexttoimage}, a masked discrete diffusion model with an
MMDiT~\citep{esser2024scalingrectifiedflowtransformers} backbone, and Janus-1.3B~\citep{wu2024janusdecouplingvisualencoding}, an autoregressive
model with a LLaMA-style~\citep{touvron2023llamaopenefficientfoundation} decoder.

\begin{algorithm}[!htb]
\caption{Head-level condition-change awareness probing.}
\label{alg:probing}
\begin{algorithmic}[1]
\State \textbf{Input:} prompt set $\mathcal{P}$; generation steps $\mathcal{T}$
\State \textbf{Output:} condition-change awareness scores $\delta^{(l,h)}(p,t)$ for every head $(l,h)$
\For{$p \in \mathcal{P}$}
  \State run one CFG generation, hooking every head's activation on both the positive and negative branches
  \For{$t \in \mathcal{T}$}
    \State $\delta^{(l,h)}(p,t) \gets \lVert \mathbf{h}_{\mathrm{pos}}^{(l,h),t} - \mathbf{h}_{\mathrm{neg}}^{(l,h),t} \rVert_2 \quad \forall (l,h)$
  \EndFor
\EndFor
\State \Return $\{\delta^{(l,h)}(p,t)\}$ \Comment{diffusion: kept per step; AR: averaged over $t$}
\end{algorithmic}
\end{algorithm}

\begin{figure}[!htb]
\centering
\includegraphics[width=0.9\linewidth]{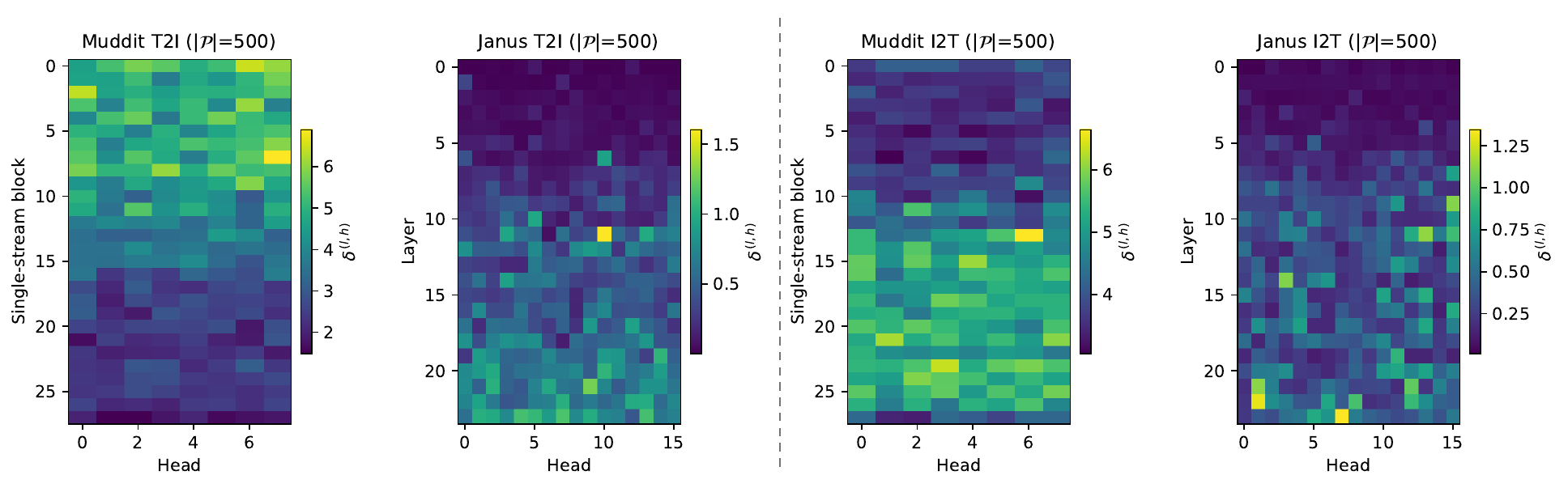}
\caption{Layer-by-head condition-change awareness $\delta^{(l,h)}$ of Muddit and Janus, averaged over $|\mathcal{P}|=500$ probing samples
(and over steps for Muddit). Left of the dashed line: T2I generation; right: I2T generation.}
\label{fig:heatmaps}
\end{figure}

Let $\mathcal{P}$ denote the set of probing samples, i.e., prompts for T2I and images for I2T, and $\mathcal{T}$ the forward passes of one generation, i.e., the denoising steps for diffusion models and the generation steps for AR models.
At each step, the model performs a standard CFG forward pass, and we hook $\mathbf{h}^{(l,h)}$, an intermediate activation of
each attention head $(l,h)$, i.e., the $h$-th head of the $l$-th layer, on both branches; the exact activation and its positions depend on the model architecture and are given in
Appendix~\ref{app:probing-impl}. We then compute $\delta^{(l,h)} = \lVert \mathbf{h}_{\mathrm{pos}}^{(l,h)} - \mathbf{h}_{\mathrm{neg}}^{(l,h)} \rVert_2$
as its \emph{condition-change awareness}, the response of the head to the change of the condition between the two branches.
Algorithm~\ref{alg:probing} gives the probing procedure. We apply this probing to the T2I and I2T generation of both models with $|\mathcal{P}|=500$ samples per task (settings in
Appendix~\ref{app:probing-impl}), and average $\delta^{(l,h)}$ over $\mathcal{P}$, and over steps for diffusion models, to obtain
the layer-by-head awareness map of each model and task in Figure~\ref{fig:heatmaps}.

\subsection{Text-to-Image Generation}
\label{sec:probing}
We first probe T2I generation on 500 GenEval~\citep{ghosh2023genevalobjectfocusedframeworkevaluating} prompts for both models
(Appendix~\ref{app:probing-t2i}).

\textbf{Different Heads Exhibit Heterogeneous Condition-change Awareness.}
We quantify the condition-change awareness of each head per prompt by $\delta^{(l,h)}$ averaged over generation steps.
As shown in Figure~\ref{fig:heatmaps} (left), both Muddit and Janus exhibit substantial variation across heads: some heads respond strongly to the condition change, whereas others barely respond.

\textbf{Head-level Condition-change Awareness Is Stable Across Prompts.}
We next ask whether these head-level differences reflect persistent head behavior or are specific to individual prompts.
We compute the mean pairwise Spearman rank correlation between per-prompt head rankings over $|\mathcal{P}|=500$ prompts.
For Janus T2I, each prompt yields a single ranking averaged over decode positions, with a correlation of $0.970 \pm 0.014$.
For Muddit T2I, rankings are computed independently at each denoising step; the correlation averages $0.975$ across steps and remains above $0.97$ except for the final few steps (Figure~\ref{fig:muddit-spearman}, left, in Appendix~\ref{app:spearman-muddit}).
These results indicate that a head's relative awareness of condition changes is largely intrinsic rather than prompt-specific, providing a stable signal for characterizing head behavior.

\textbf{Condition-change Aware Heads Attend to Semantically Relevant Regions.}
Figure~\ref{fig:teaser} reveals a clear functional distinction between condition-change aware and condition-change agnostic heads.
We select heads from the high end of the probed $\delta^{(l,h)}$ as condition-change aware and from the low end as condition-change agnostic, and visualize the attention of each generated image token to words in the prompt.
Aware heads concentrate their attention on the region of the queried keyword, whereas agnostic heads attend diffusely with little semantic correspondence.

\subsection{Image-to-Text Generation}
\label{sec:probing-i2t}
For I2T generation, CFG is not a standard component of decoding.
However, methods such as
Visual Contrastive Decoding (VCD)~\citep{leng2023mitigatingobjecthallucinationslarge} use a specially constructed perturbed image as the negative branch of CFG and improve the robustness of VLMs against hallucination.
Our probing analysis follows this construction, taking the perturbed image as the negative condition.
This is conducted on the captioning task over
500 MS-COCO~\citep{lin2015microsoftcococommonobjects} images (Appendix~\ref{app:probing-i2t}).

\textbf{Results.} I2T generation exhibits the same heterogeneity as T2I generation: on both models a subset of heads responds strongly to the
change of the image while the rest barely respond (Figure~\ref{fig:heatmaps}, right).
Similarly, Figure~\ref{fig:teaser} (right) shows the distinct pattern on captioning, where the aware head attends to the object relevant to the decoded keyword.
The ranking of Janus is stable across samples, with a mean pairwise Spearman rank correlation of $0.962$ (Table~\ref{tab:janus-spearman});
However, the ranking of Muddit degrades steadily over the 32 denoising steps, staying above $0.90$ only in the first 9 steps and
falling to $0.67$ at the last (Figure~\ref{fig:muddit-spearman}, right, in Appendix~\ref{app:spearman-muddit}).

\section{Method: Tuning CFG at Head-level}
\label{sec:method}
This section presents HeadTuned-CFG, which tunes CFG at the head level according to the probed condition-change awareness in two
steps: selecting heads by their awareness and adjusting the guidance strength of the selected heads.

\begin{algorithm}[!b]
\caption{HeadTuned-CFG (\textit{select} and \textit{tune}).}
\label{alg:headtune}
\begin{algorithmic}[1]
\State \textbf{Input:} scores $\delta^{(l,h)}$; mode (interpolation: bottom-$k$ heads, $s<1$; extrapolation: top-$k$ heads, $s>1$); $\Phi_s$; $w$; steps $\mathcal{T}$, tuning steps $\mathcal{T}' \subseteq \mathcal{T}$
\State \textbf{Output:} generated sample
\For{$t \in \mathcal{T}$}
  \If{$t \in \mathcal{T}'$}
    \State $\mathcal{S}_k \gets$ the $k$ heads of the mode by $\delta^{(l,h)}$ \Comment{per step for diffusion; fixed for AR}
    \State run the positive and negative branches, replacing in the forward pass
    \Statex \hspace{\algorithmicindent}\hspace{\algorithmicindent} $\mathbf{h}_{\mathrm{pos}}^{(l,h)} \gets \Phi_s\!\left(\mathbf{h}_{\mathrm{pos}}^{(l,h)}, \mathbf{h}_{\mathrm{neg}}^{(l,h)}\right) \quad \forall (l,h) \in \mathcal{S}_k$
  \Else
    \State run the positive and negative branches
  \EndIf
  \State combine the two branches with guidance scale $w$ \Comment{Eq.~\ref{eq:cfg}}
\EndFor
\State \Return sample
\end{algorithmic}
\end{algorithm}

\textbf{Motivation.} Section~\ref{sec:awareness} reveals functional specialization among attention heads under CFG.
We hypothesize that the response of a condition-change agnostic head to the change of condition carries little
condition-specific information and acts as noise, which CFG nevertheless amplifies together with the informative
responses, whereas the response of a condition-change aware head points in the direction of the condition itself.
This suggests adjusting the guidance per head rather than on the final output alone: interpolating the positive-branch
activation of agnostic heads toward the negative branch suppresses the amplified noise, and
extrapolating that of aware heads away from the negative branch reinforces the condition-specific direction and
strengthens condition adherence.
To verify this conjecture, we provide a theoretical justification in a simplified setting under a local Gaussian model in
Theorem~\ref{thm:headwise-kl-projection}, with the proof deferred to
Appendix~\ref{app:headtuning-theory}.

\begin{theorem}
\label{thm:headwise-kl-projection}
For one head, suppress $(l,h)$ and let
$p_b=\mathcal{N}(\boldsymbol{\mu}_b,\Sigma)$ with $\Sigma \succ 0$ for $b\in\{\mathrm{pos},\mathrm{neg}\}$.
Define LDA direction
$\mathbf{w}=\Sigma^{-1}
(\boldsymbol{\mu}_{\mathrm{pos}}-\boldsymbol{\mu}_{\mathrm{neg}})$ with $\mu_{\mathrm{pos}} \ne \mu_{\mathrm{neg}}$,
and induced positive-awareness score $T(\mathbf{x})=\mathbf{w}^{\top}\mathbf{x}$.
The KL projection
\begin{equation}
    \min_q \operatorname{KL}(q\|p_{\mathrm{neg}})
    \quad\text{s.t.}\quad
    \mathbb{E}_q[T(\mathbf{x})]\geq\tau
    \label{eq:headwise-kl-projection}
\end{equation}
has solution $q^*=\mathcal{N}(\mathbf{h}_g,\Sigma)$,
whose maximum a posteriori (MAP) estimate and tuning strength are
\begin{equation}
    \mathbf{h}_g
    =\boldsymbol{\mu}_{\mathrm{neg}}
    +s^*(\boldsymbol{\mu}_{\mathrm{pos}}
         -\boldsymbol{\mu}_{\mathrm{neg}}),
    \qquad
    s^*=
    \left[
        \frac{\tau-\mathbf{w}^{\top}
        \boldsymbol{\mu}_{\mathrm{neg}}}
        {\mathbf{w}^{\top}\Sigma\mathbf{w}}
    \right]_+.
    \label{eq:headwise-optimal-strength}
\end{equation}
Moreover, $s^*>1$ (resp., $s^*<1$) if and only if
$\tau>\mathbb{E}_{p_{\mathrm{pos}}}[T(\mathbf{x})]$
(resp., $\tau<\mathbb{E}_{p_{\mathrm{pos}}}[T(\mathbf{x})]$).
\end{theorem}

Theorem~\ref{thm:headwise-kl-projection} interprets head-wise tuning
as a KL projection: we seek the distribution closest to the negative
branch that attains a prescribed positive-awareness level.
Under the local Gaussian model, its most probable activation is
obtained by \textit{interpolating or extrapolating} between the two branch
means. Replacing these means with observed activations recovers our
tuning rule. Targets above or below the unchanged positive branch's
awareness level yield extrapolation ($s>1$) or interpolation ($s<1$),
respectively. This motivates the two modes in our HeadTuned-CFG algorithm (see Algorithm~\ref{alg:headtune}): (a) extrapolating
the top-$k$ heads ranked by condition-change awareness to strengthen
conditioning, and (b) interpolating the bottom-$k$ heads toward the
negative branch to suppress their uninformative response to the condition. Each mode uses a shared
tuning strength $s$ across the selected heads.

\begin{figure}[!b]
\centering
\includegraphics[width=0.80\linewidth]{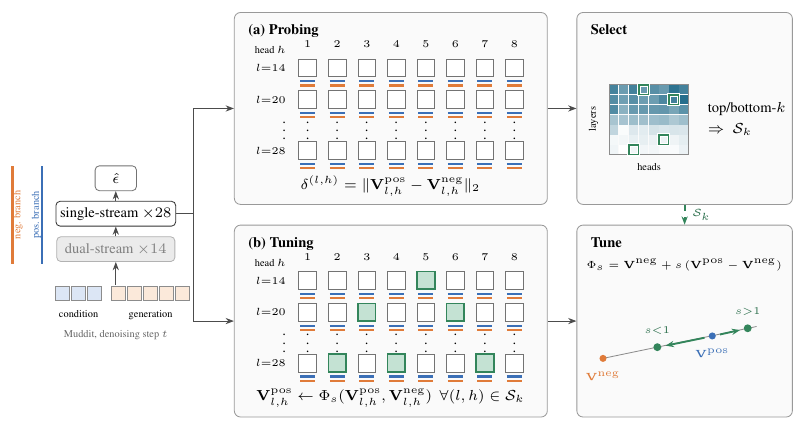}
\caption{Proposed HeadTuned-CFG method on Muddit. Illustration of head-level probing, selection, and tuning process of the single-stream attention heads.}
\label{fig:muddit-arch}
\end{figure}

\textbf{Algorithm.} Given the condition-change awareness scores $\delta^{(l,h)}$, HeadTuned-CFG proceeds in two steps, \emph{select} and \emph{tune}
(Algorithm~\ref{alg:headtune}), controlled by two hyperparameters: the fraction $k$ of heads to tune and the tuning strength $s$.

\emph{Select}: heads are ranked by $\delta^{(l,h)}$, and the bottom-$k$ fraction (interpolation mode) or the top-$k$ fraction (extrapolation
mode) forms the tuning set $\mathcal{S}_k$; for masked-diffusion models the set is selected at each denoising step from the per-step scores, and for
AR models once from the averaged scores.

\emph{Tune}: at a subset of steps $\mathcal{T}' \subseteq \mathcal{T}$, a tuning operator $\Phi_s$ is applied to the selected heads in
$\mathcal{S}_k$ and modifies their positive-branch activation, displacing it along the direction between the positive and negative branches:
$s=1$ leaves the head unchanged and recovers vanilla CFG, $s<1$ interpolates it toward the negative branch, and $s>1$ extrapolates it away
from the negative branch.
For simplicity, the same $s$ is applied uniformly to all selected heads.
The interpolation mode thus tunes the bottom-$k$ heads with $s<1$ to suppress the noise amplified by CFG, whereas the extrapolation mode tunes the top-$k$
heads with $s>1$ to strengthen condition adherence.
The concrete form of $\Phi_s$ and the choice of $\mathcal{T}'$ depend on the architecture and are given in Appendix~\ref{app:headtuning-impl}.
Figure~\ref{fig:muddit-arch} illustrates the procedure on Muddit.

\section{Experiments}
\subsection{Text-to-Image Generation}
\label{sec:t2i-exp}
\textbf{Setup.} We evaluate Muddit and Janus on MJHQ-30K~\citep{li2024playgroundv25insightsenhancing}, which contains 30{,}000 prompts in 10
categories of 3{,}000 prompts each. FID~\citep{heusel2018ganstrainedtimescaleupdate} is the primary metric.
Generation settings (resolution, denoising steps, and the negative branch) follow the probing setup in Appendix~\ref{app:probing-t2i}, and
the head ranking is the T2I probe of Section~\ref{sec:probing}, obtained on GenEval prompts disjoint from MJHQ.
We use the interpolation mode (bottom-$k$ heads, $s<1$) here, and grid-search $(w, k, s)$ for
the lowest FID
on the \emph{animals} category (MJHQ-animals; Appendix~\ref{app:grid-t2i}), then apply the selected setting unchanged to all 10 categories
(MJHQ-full) and compare it with vanilla CFG at its best $w$.
PAG~\citep{ahn2025selfrectifyingdiffusionsamplingperturbedattention}, a guidance method that is likewise designed inside the architecture and
can be stacked on CFG, is included as an additional baseline on Muddit, alone and on top of CFG, with the hyperparameters recommended in its
paper.
Tables~\ref{tab:t2i-muddit} and \ref{tab:t2i-janus} report the FID per category and on MJHQ-full; Appendix~\ref{app:full-metrics} describes
and reports the full set of other scores.

\textbf{Results.} On Muddit, HeadTuned-CFG lowers the FID on MJHQ-full from $20.90$ to $16.29$ and the FID of every category
(Table~\ref{tab:t2i-muddit}); on Janus it lowers the FID on MJHQ-full from $10.51$ to $9.92$ and the FID of 9 of the 10 categories
(Table~\ref{tab:t2i-janus}).
PAG alone improves over a single forward pass on Muddit but falls short of CFG, and stacking it on CFG degrades CFG (Table~\ref{tab:t2i-muddit}).
The full metrics in Appendix~\ref{app:full-metrics} show that the lower FID is not obtained at the expense of the other scores.
In short, HeadTuned-CFG outperforms the related guidance methods on the T2I generation of two models of different paradigms and
architectures.

\begin{table}[H]
\centering
\footnotesize
\setlength{\tabcolsep}{2.4pt}
\caption{Muddit T2I: FID per MJHQ category ($n=3{,}000$ each) and on MJHQ-full ($n=30{,}000$); CFG ($w=3$), PAG ($\lambda=2.5$, single-stream
blocks 9--10), PAG+CFG ($\lambda=1.5$ on CFG with $w=3$), and HeadTuned-CFG ($w=9$, $k=50\%$, $s=0.4$).}
\label{tab:t2i-muddit}
\renewcommand{\arraystretch}{0.92}
\begin{tabular}{l *{10}{c} @{\hspace{5pt}} c}
\toprule
& animals & art & fashion & food & indoor & landscape & logo & people & plants & vehicles & full \\
\midrule
CFG & 33.91 & 42.82 & 38.79 & 39.53 & 43.82 & 40.93 & 40.27 & 43.85 & 49.93 & 36.89 & 20.90 \\
PAG & 52.31 & 57.31 & 53.30 & 64.44 & 61.24 & 65.27 & 51.48 & 60.09 & 76.86 & 54.04 & 34.37 \\
PAG+CFG & 38.21 & 44.40 & 40.03 & 42.27 & 43.74 & 45.92 & 39.46 & 45.28 & 56.44 & 38.59 & 22.39 \\
HeadTuned-CFG & \textbf{30.41} & \textbf{36.70} & \textbf{34.02} & \textbf{33.60} & \textbf{36.17} & \textbf{34.56} & \textbf{36.65} & \textbf{36.88} & \textbf{43.01} & \textbf{30.93} & \textbf{16.29} \\
\bottomrule
\end{tabular}
\vspace{10pt}
\centering
\footnotesize
\setlength{\tabcolsep}{2.4pt}
\caption{Janus T2I: FID per MJHQ category ($n=3{,}000$ each) and on MJHQ-full ($n=30{,}000$). CFG ($w=4$) versus HeadTuned-CFG
($w=4$, $k=15\%$, $s=0.75$); PAG is not applicable to Janus.}
\label{tab:t2i-janus}
\begin{tabular}{l *{10}{c} @{\hspace{5pt}} c}
\toprule
& animals & art & fashion & food & indoor & landscape & logo & people & plants & vehicles & full \\
\midrule
CFG & 27.32 & 31.33 & 30.10 & \textbf{31.05} & 28.06 & 27.53 & 37.90 & 33.97 & 30.20 & 23.66 & 10.51 \\
HeadTuned-CFG & \textbf{25.95} & \textbf{29.93} & \textbf{28.83} & 31.37 & \textbf{27.03} & \textbf{24.81} & \textbf{37.60} & \textbf{32.02} & \textbf{29.37} & \textbf{23.16} & \textbf{9.92} \\
\bottomrule
\end{tabular}
\end{table}

\subsection{Ablation Study on T2I Head Selection}
\label{sec:ablation}
\begin{figure}[!htb]
\centering
\includegraphics[width=0.68\linewidth]{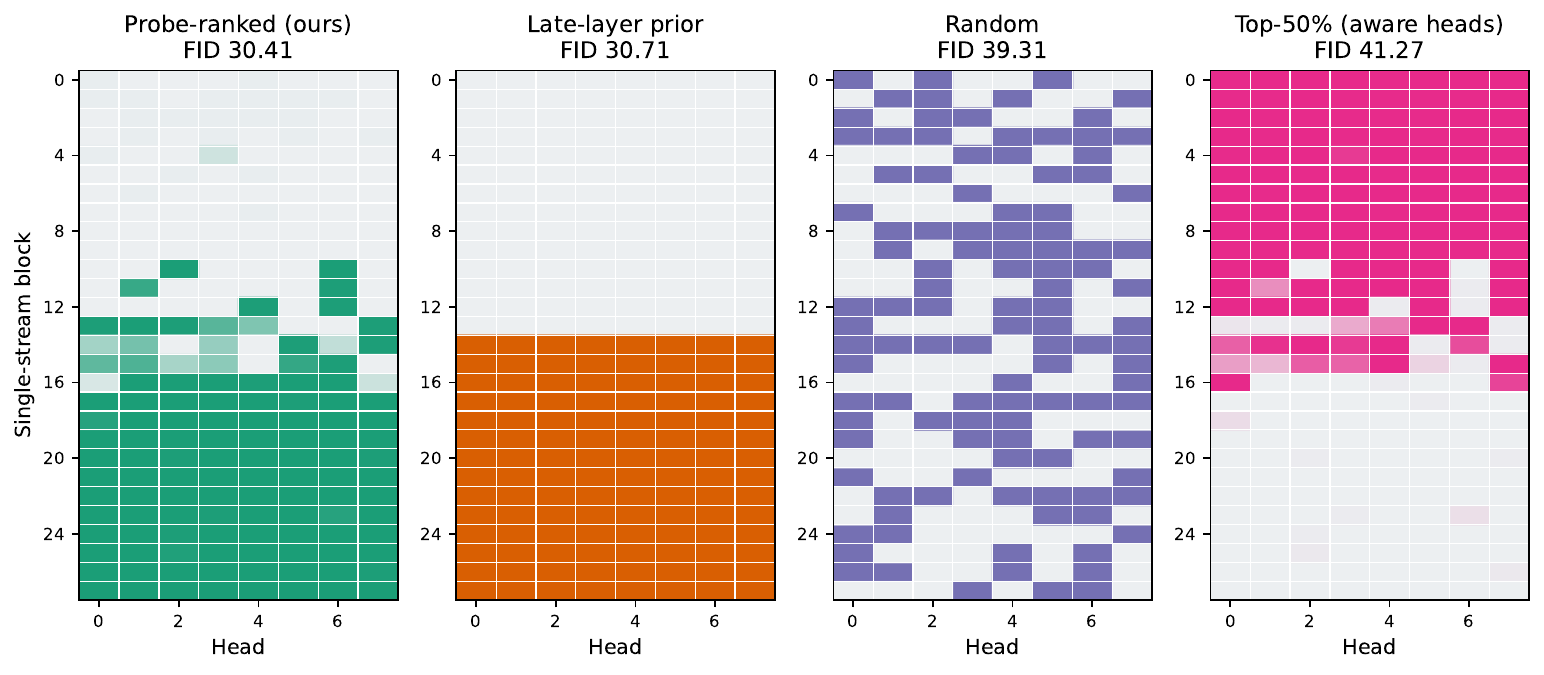}
\caption{Heads tuned by the four selection rules on Muddit (single-stream block by head), with FID on MJHQ-animals at $w=9$, $k=50\%$,
$s=0.4$. The probed and the top-$50\%$ selections change per denoising step; their panels show the fraction of the 64 steps in
which a head is tuned.}
\label{fig:headsel-muddit}
\end{figure}

To test whether the gain comes from the probed ranking, on Muddit we keep all parameters $w=9$, $k=50\%$, $s=0.4$, while only replacing the probe-based selection
by a late-layer prior (all heads of the last 14 single-stream blocks, where most of the probed bottom-$50\%$ heads lie), a random set of
the same size, and the reverse selection that tunes the top-$50\%$ heads, i.e., the condition-change aware heads
(Figure~\ref{fig:headsel-muddit}).
On MJHQ-animals, the FID ranks probe-based ($30.41$) $<$ late-layer prior ($30.71$) $<$ random ($39.31$) $<$ reverse ($41.27$),
against $33.91$ for vanilla CFG: the probe-based selection is better than the layer prior, whereas random and reverse selection are both
worse than no tuning, which confirms that the probed condition-change agnostic heads are the right heads to tune.
The same ablation on Janus gives the same ordering (Appendix~\ref{app:headsel}).

\subsection{Image-to-Text Generation}
\label{sec:i2t-exp}

\textbf{Setup.} We evaluate object hallucination of Janus on POPE~\citep{li2023evaluatingobjecthallucinationlarge} and on the hallucination
subset of MME~\citep{fu2025mmecomprehensiveevaluationbenchmark}.
POPE asks yes/no questions about the presence of objects in MS-COCO images on three splits (\emph{random}, \emph{popular}, and
\emph{adversarial}; 3{,}000 questions each) and reports accuracy and F1; the hallucination subset of MME covers \emph{existence},
\emph{count}, \emph{position}, and \emph{color}, scored as in VCD~\citep{leng2023mitigatingobjecthallucinationslarge} out of 200 per subtask.
We follow the protocol and the decoding settings of VCD (Appendix~\ref{app:grid-i2t}); the head ranking is the I2T probe of
Section~\ref{sec:probing-i2t}, obtained on the captioning task.
\emph{Regular} denotes decoding without guidance, \emph{VCD} the decoding above, and \emph{HeadTuned-VCD} adds head tuning to VCD in the
extrapolation mode (top-$k$ heads, $s>1$) to further strengthen the output of the condition-change aware heads.
As for T2I, we sweep $w$ for VCD and grid-search $(w, k, s)$ for HeadTuned-VCD and report the best setting of each
(Appendix~\ref{app:grid-i2t}); on POPE the search uses the random split, and the selected setting is applied unchanged to the other two
splits.

\begin{table}[!htb]
\centering
\footnotesize
\caption{Janus on POPE, accuracy and F1 (\%) per split ($n=3{,}000$ each): Regular decoding, VCD ($w=8$, $\beta=0.2$, $T=500$), and
HeadTuned-VCD ($w=8$, $k=20$ heads, $s=1.3$).}
\label{tab:janus-pope}
\begin{tabular}{l cc cc cc cc}
\toprule
& \multicolumn{2}{c}{random} & \multicolumn{2}{c}{popular} & \multicolumn{2}{c}{adversarial} & \multicolumn{2}{c}{overall} \\
& Acc & F1 & Acc & F1 & Acc & F1 & Acc & F1 \\
\midrule
Regular & 84.17 & 81.60 & 82.97 & 80.47 & 81.33 & 78.99 & 82.82 & 80.35 \\
VCD & 87.77 & 86.33 & 85.97 & 84.63 & 83.93 & 82.79 & 85.89 & 84.58 \\
HeadTuned-VCD & \textbf{88.80} & \textbf{87.76} & \textbf{86.43} & \textbf{85.54} & \textbf{84.13} & \textbf{83.50} & \textbf{86.45} & \textbf{85.60} \\
\bottomrule
\end{tabular}
\end{table}

\begin{table}[!htb]
\centering
\footnotesize
\caption{Janus on the hallucination subset of MME, score (accuracy $+$ image-level accuracy, out of 200) per subtask and in total:
Regular decoding, VCD ($w=10$, $\beta=0.2$, $T=500$), and HeadTuned-VCD ($w=8$, $k=15$ heads, $s=1.1$).}
\label{tab:janus-mme}
\begin{tabular}{l cccc c}
\toprule
& existence & count & position & color & total \\
\midrule
Regular & 138.33 & 56.67 & 90.00 & 93.33 & 378.33 \\
VCD & \textbf{170.00} & 81.67 & \textbf{91.67} & 123.33 & 466.67 \\
HeadTuned-VCD & \textbf{170.00} & \textbf{83.33} & \textbf{91.67} & \textbf{130.00} & \textbf{475.00} \\
\bottomrule
\end{tabular}
\end{table}

\textbf{Results.} On POPE, HeadTuned-VCD raises the accuracy of the random split from $87.77$ to $88.80$ and the F1 from $86.33$ to $87.76$
over VCD, and improves on every split in both metrics when applied unchanged to the other splits (Table~\ref{tab:janus-pope}).
On the hallucination subset of MME, VCD raises the total score from $378.33$ to $466.67$, with the largest gains on existence and color;
HeadTuned-VCD further raises it to $475.00$, matching VCD on existence and position and improving count and color (Table~\ref{tab:janus-mme}).

\textbf{Limitation.} We also applied HeadTuned-VCD to Muddit on the same benchmarks, but it did not improve over VCD.
Several factors contribute: Muddit itself performs poorly on POPE and MME, and VCD brings only a marginal gain on it.
Moreover, as discussed in Section~\ref{sec:probing-i2t}, its condition-change awareness ranking in I2T generation is unstable across samples in the
late denoising steps, which makes the selected heads unreliable across samples.
We discuss this limitation in I2T generation here, as it may motivate further study in the future.

\section{Conclusion and Future Work}
\label{sec:conclusion}
In this work, we investigate the shared backbone of unified VLMs through the lens of conditional guidance.
Probing the intermediate activations of attention heads across the two branches of CFG reveals that the heads are functionally
specialized: a subset of them is aware of the change of condition, while some are agnostic to it.
This specialization appears consistently in both T2I and I2T generation and is largely stable across samples, suggesting that it is an
intrinsic property of unified VLMs.
Building on this specialization, HeadTuned-CFG uses the probed condition-change awareness to select heads and tune the guidance at the head level, and
brings consistent benefits to both generation directions of two unified VLMs of different paradigms, including substantial improvements
over vanilla CFG and related guidance methods in T2I generation.
These results in turn support the probing analysis, and establish conditional guidance as a new perspective for understanding unified VLMs.

Two directions follow from this work.
First, HeadTuned-CFG currently selects its hyperparameters $(w, k, s)$ by grid search on a validation subset; a more efficient tuning,
e.g., setting the head-wise strength from the probed condition-change awareness itself as suggested by the analysis in
Appendix~\ref{app:headtuning-theory}, would make the method less costly to apply to new models and tasks.
Second, the head-level specialization revealed by our probing is a property of the pretrained backbone, which raises an intriguing question of how it
emerges.
Understanding this could inform the architecture design and the training paradigm of unified VLMs, for instance by encouraging or
exploiting such specialization when both T2I and I2T generation are trained jointly.

\section{Acknowledgment}
LS acknowledges funding support by National Science Foundation (NSF) via grant IIS-2435746, Defense Advanced Research Projects Agency (DARPA) under contract No. HR00112520042, as well as the University of Michigan MICDE Catalyst Grant Award and MIDAS PODS Grant Award. ZT and Yingbin L acknowledge funding support by U.S.\ National Science Foundation under the grants MFAI-2601957 and ECCS-2515482.  

\bibliographystyle{plainnat}
\bibliography{references}

\clearpage
\appendix
\raggedbottom
\section{Visualization of Condition-change Aware and Agnostic Heads on Muddit}
\label{app:teaser-muddit}

\begin{figure}[H]
\centering
\includegraphics[width=\linewidth]{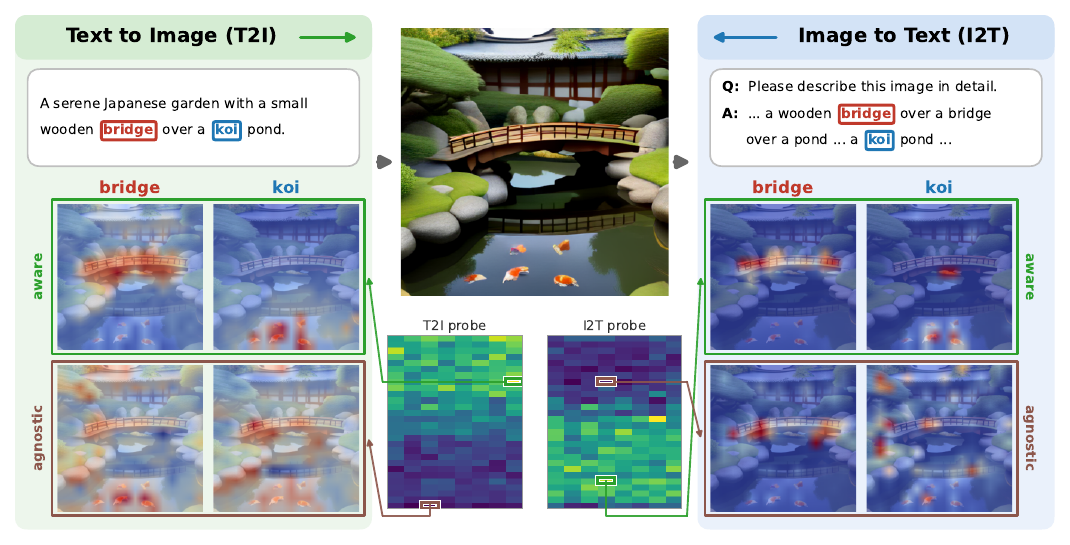}
\caption{Condition-change aware (green) and agnostic (brown) attention heads of Muddit, in the layout of Figure~\ref{fig:teaser}: the
T2I maps show the alignment of each head's CFG difference with the keyword in value space, the I2T maps the attention of the decoded keyword
to the image, and the middle column the generated image and the Muddit probes with the visualized heads marked.}
\label{fig:teaser-muddit}
\end{figure}

Figure~\ref{fig:teaser-muddit} repeats Figure~\ref{fig:teaser} on Muddit with the same prompt and the heads selected by the Muddit probes.
The T2I maps show where the CFG difference of the head aligns with the keyword in value space, and the I2T maps show the attention of the
decoded keyword to the image, as for Janus.

\section{Implementation Details of Probing}
\label{app:probing-impl}

\textbf{Probing procedure.} Probing runs standard CFG generation with the positive and negative branches in one batch and hooks the activations
of every head without modifying them.
At a hooked head $(l,h)$ and step $t$, the score $\delta^{(l,h)}(p,t)$ in Algorithm~\ref{alg:probing} is the $\ell_2$ norm of
$\mathbf{h}_{\mathrm{pos}}^{(l,h),t} - \mathbf{h}_{\mathrm{neg}}^{(l,h),t}$ over the head dimension $\bar{d}$, computed at the generation positions
and averaged over them.
In Muddit the hook reads the per-head value vectors $\mathbf{V}_{l,h}$, and in Janus the per-head attention outputs $\mathbf{O}_{l,h}$ (Section~\ref{sec:mha}).
The prompt set $\mathcal{P}$ contains $|\mathcal{P}|=500$ samples, and the ranking used for selection averages $\delta^{(l,h)}$ over $\mathcal{P}$.
The construction of the two branches, the probed layers and positions, and the generation parameters are given per task and model below.

\textbf{Muddit.} Muddit uses an MMDiT backbone with 14 dual-stream and 28 single-stream blocks of 8 heads each; every denoising step processes the
full sequence in one forward pass.
The sequence concatenates the tokens of the two modalities; we call the positions of the conditioning modality the \emph{condition positions} and
those of the generated modality the \emph{generation positions}.
The 28 single-stream blocks are probed, and the ranking is over these 224 heads.
Scores are kept per denoising step, giving $\delta^{(l,h)}(p,t)$ of shape $|\mathcal{T}| \times 28 \times 8$ per sample.
Generation runs at $512 \times 512$ resolution.

\textbf{Janus.} Janus-1.3B uses a LLaMA-style decoder with 24 layers of 16 heads; the condition is encoded in a prefill pass, after which the output
is decoded autoregressively with a KV cache.
The condition positions are the prefill tokens and the generation positions the decoded tokens.
All 24 layers are probed at the decode steps; with the KV cache, each decode step computes attention outputs only for the newly generated
position, which is therefore the probed position.
Scores are averaged over the decode steps, giving $\delta^{(l,h)}(p)$ of shape $24 \times 16$ per sample.
Image generation runs at $384 \times 384$ resolution, i.e., 576 image tokens decoded in 576 steps.

\subsection{Probing on T2I Generation}
\label{app:probing-t2i}

$\mathcal{P}$ contains 500 GenEval prompts for both models.
In Muddit, the negative branch replaces the prompt with a negative prompt, and generation uses $|\mathcal{T}|=64$ and $w=9$.
In Janus, the negative branch replaces the text tokens with padding tokens, and generation uses $w=5$.

\subsection{Probing on I2T Generation}
\label{app:probing-i2t}

$\mathcal{P}$ contains 500 images of MS-COCO, and the prompt asks for a detailed description of the image.
The negative branch replaces the image with a noised copy as in VCD ($T=500$), and generation uses $w=2$ with greedy decoding.

\section{Cross-sample Stability of the Condition-change Awareness Ranking}
\label{app:spearman-muddit}

\begin{figure}[H]
\centering
\includegraphics[width=0.7\linewidth]{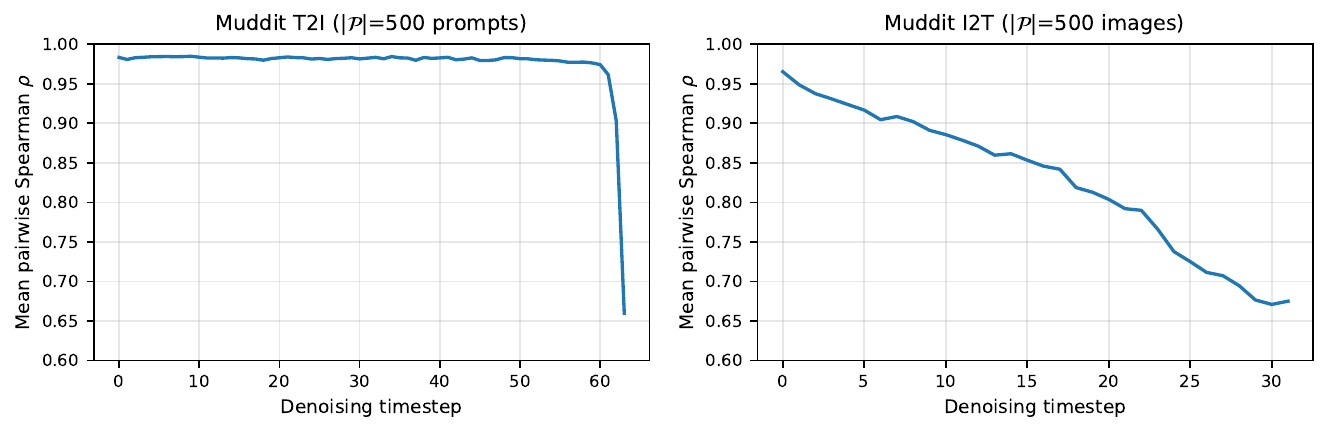}
\caption{Cross-sample stability of the condition-change awareness ranking in Muddit, per denoising step. Left: T2I generation. Right: I2T generation.}
\label{fig:muddit-spearman}
\end{figure}

Figure~\ref{fig:muddit-spearman} plots the mean pairwise Spearman rank correlation of the per-sample head rankings at each denoising step for Muddit
($|\mathcal{P}|=500$ samples per task); the I2T ranking is probed as in Appendix~\ref{app:probing-i2t} with $|\mathcal{T}|=32$ and $w=1.25$.
Table~\ref{tab:janus-spearman} reports the same statistic for Janus, whose samples each yield one ranking averaged over decode positions.
Both rankings of Janus and the T2I ranking of Muddit are stable across samples, whereas the I2T ranking of Muddit degrades over the
denoising steps, from $0.97$ at the first step to $0.67$ at the last.

\begin{table}[H]
\centering
\small
\caption{Mean pairwise Spearman rank correlation of the condition-change awareness ranking of Janus across $|\mathcal{P}|=500$ probing samples.}
\label{tab:janus-spearman}
\begin{tabular}{l cc}
\toprule
& T2I & I2T \\
\midrule
Spearman $\rho$ & $0.970 \pm 0.014$ & $0.962 \pm 0.023$ \\
\bottomrule
\end{tabular}
\end{table}

\section{Theoretical Justification of HeadTuned-CFG}
\label{app:headtuning-theory}

This section proves Theorem~\ref{thm:headwise-kl-projection} and gives a theoretical justification of the head-wise update in HeadTuned-CFG.
We show that the tuned head activation is the MAP estimate of the distribution obtained by a KL projection from the negative branch under a
positive-awareness constraint.
This view also explains the two modes of Section~\ref{sec:method}: the probed ranking separates condition-change aware heads from
condition-change agnostic heads, and the two groups receive different positive-awareness targets, so that the former amplify condition-specific
information ($s>1$) while the latter retain only a fraction of their uninformative response and stay close to the negative branch ($s<1$).

Consider one probed head $(l,h)$ with positive- and negative-branch activations $\mathbf{h}_{\mathrm{pos}}^{(l,h)}$ and
$\mathbf{h}_{\mathrm{neg}}^{(l,h)}$. HeadTuned-CFG modifies the positive branch by the tuning operator
\[
\Phi_s\!\left(\mathbf{h}_{\mathrm{pos}}^{(l,h)}, \mathbf{h}_{\mathrm{neg}}^{(l,h)}\right)
= \mathbf{h}_{\mathrm{neg}}^{(l,h)} + s\left(\mathbf{h}_{\mathrm{pos}}^{(l,h)} - \mathbf{h}_{\mathrm{neg}}^{(l,h)}\right).
\]

\textbf{Head-wise CFG model.}
For head $(l,h)$, suppose the activations of the two branches locally follow two Gaussian distributions with a shared covariance,
\[
\mathbf{x}_{\mathrm{pos}}^{(l,h)} \sim p_{\mathrm{pos}}^{(l,h)} := \mathcal{N}\!\left(\boldsymbol{\mu}_{\mathrm{pos}}^{(l,h)}, \Sigma_{l,h}\right),
\qquad
\mathbf{x}_{\mathrm{neg}}^{(l,h)} \sim p_{\mathrm{neg}}^{(l,h)} := \mathcal{N}\!\left(\boldsymbol{\mu}_{\mathrm{neg}}^{(l,h)}, \Sigma_{l,h}\right).
\]
Linear discriminant analysis (LDA) gives the discriminant direction
\[
\mathbf{w}_{l,h} = \Sigma_{l,h}^{-1}\left(\boldsymbol{\mu}_{\mathrm{pos}}^{(l,h)} - \boldsymbol{\mu}_{\mathrm{neg}}^{(l,h)}\right),
\]
and we call $T_{l,h}(\mathbf{x}) = \mathbf{w}_{l,h}^{\top}\mathbf{x}$ the \emph{positive-awareness score} of an activation $\mathbf{x}$ at this head.

We then seek the distribution $q_{l,h}$ closest to the negative-branch distribution $p_{\mathrm{neg}}^{(l,h)}$ that attains a target
positive-awareness level $\tau_{l,h}$:
\begin{equation}
\label{eq:primal}
\min_{q_{l,h}} \mathrm{KL}\!\left(q_{l,h} \,\|\, p_{\mathrm{neg}}^{(l,h)}\right)
\quad \text{s.t.} \quad
\mathbb{E}_{q_{l,h}}\!\left[T_{l,h}(\mathbf{x})\right] \ge \tau_{l,h}.
\end{equation}
The Lagrangian is
\[
\mathcal{L}(q_{l,h}, s_{l,h}) = \mathrm{KL}\!\left(q_{l,h} \,\|\, p_{\mathrm{neg}}^{(l,h)}\right)
+ s_{l,h}\left(\tau_{l,h} - \mathbb{E}_{q_{l,h}}\!\left[T_{l,h}(\mathbf{x})\right]\right),
\]
with multiplier $s_{l,h} \ge 0$. For fixed $s_{l,h}$, minimizing over $q_{l,h}$ gives the exponential tilt
\[
q^{*}_{l,h,s_{l,h}}(\mathbf{x}) \propto p_{\mathrm{neg}}^{(l,h)}(\mathbf{x}) \exp\!\left(s_{l,h} T_{l,h}(\mathbf{x})\right).
\]
Substituting it back yields the dual objective
\[
g_{l,h}(s_{l,h}) = s_{l,h}\tau_{l,h} - \log Z_{l,h}(s_{l,h}),
\]
where
\begin{equation}
\label{eq:Z-partition}
Z_{l,h}(s_{l,h}) = \mathbb{E}_{p_{\mathrm{neg}}^{(l,h)}}\!\left[\exp\!\left(s_{l,h} T_{l,h}(\mathbf{x})\right)\right]
= \exp\!\left(s_{l,h}\,\mathbf{w}_{l,h}^{\top}\boldsymbol{\mu}_{\mathrm{neg}}^{(l,h)}
+ \frac{s_{l,h}^{2}}{2}\,\mathbf{w}_{l,h}^{\top}\Sigma_{l,h}\mathbf{w}_{l,h}\right).
\end{equation}
The last equality holds because, under $p_{\mathrm{neg}}^{(l,h)}$, $T_{l,h}(\mathbf{x}) = \mathbf{w}_{l,h}^{\top}\mathbf{x}$ is a one-dimensional
Gaussian with moment-generating function $\mathbb{E}[\exp(sT)] = \exp\!\left(s\,\mathbb{E}[T] + s^{2}\mathrm{Var}(T)/2\right)$.
The optimal multiplier solves the one-dimensional dual problem $s^{*}_{l,h} = \arg\max_{s_{l,h}\ge 0} g_{l,h}(s_{l,h})$; with
Eq.~\ref{eq:Z-partition},
\[
g_{l,h}(s_{l,h}) = s_{l,h}\left(\tau_{l,h} - \mathbf{w}_{l,h}^{\top}\boldsymbol{\mu}_{\mathrm{neg}}^{(l,h)}\right)
- \frac{s_{l,h}^{2}}{2}\,\mathbf{w}_{l,h}^{\top}\Sigma_{l,h}\mathbf{w}_{l,h},
\]
and maximizing this concave quadratic gives
\begin{equation}
\label{eq:optimal-s}
s^{*}_{l,h}
= \left[\frac{\tau_{l,h} - \mathbf{w}_{l,h}^{\top}\boldsymbol{\mu}_{\mathrm{neg}}^{(l,h)}}{\mathbf{w}_{l,h}^{\top}\Sigma_{l,h}\mathbf{w}_{l,h}}\right]_{+}
= \left[\frac{\tau_{l,h} - \left(\boldsymbol{\mu}_{\mathrm{pos}}^{(l,h)} - \boldsymbol{\mu}_{\mathrm{neg}}^{(l,h)}\right)^{\top}\Sigma_{l,h}^{-1}\boldsymbol{\mu}_{\mathrm{neg}}^{(l,h)}}
{\left(\boldsymbol{\mu}_{\mathrm{pos}}^{(l,h)} - \boldsymbol{\mu}_{\mathrm{neg}}^{(l,h)}\right)^{\top}\Sigma_{l,h}^{-1}\left(\boldsymbol{\mu}_{\mathrm{pos}}^{(l,h)} - \boldsymbol{\mu}_{\mathrm{neg}}^{(l,h)}\right)}\right]_{+}.
\end{equation}
Therefore the solution of the primal problem in Eq.~\ref{eq:primal} is
\[
q^{*}_{l,h,s^{*}_{l,h}} = \mathcal{N}\!\left(\boldsymbol{\mu}_{\mathrm{neg}}^{(l,h)} + s^{*}_{l,h}\,\Sigma_{l,h}\mathbf{w}_{l,h},\ \Sigma_{l,h}\right),
\]
whose MAP estimate is
\[
\mathbf{h}_{g}^{(l,h)} = \boldsymbol{\mu}_{\mathrm{neg}}^{(l,h)} + s^{*}_{l,h}\,\Sigma_{l,h}\mathbf{w}_{l,h}
= \boldsymbol{\mu}_{\mathrm{neg}}^{(l,h)} + s^{*}_{l,h}\left(\boldsymbol{\mu}_{\mathrm{pos}}^{(l,h)} - \boldsymbol{\mu}_{\mathrm{neg}}^{(l,h)}\right).
\]
Estimating the class means by the observed head activations, $\boldsymbol{\mu}_{\mathrm{pos}}^{(l,h)} \approx \mathbf{h}_{\mathrm{pos}}^{(l,h)}$ and
$\boldsymbol{\mu}_{\mathrm{neg}}^{(l,h)} \approx \mathbf{h}_{\mathrm{neg}}^{(l,h)}$, yields
\[
\mathbf{h}_{g}^{(l,h)} = \mathbf{h}_{\mathrm{neg}}^{(l,h)} + s^{*}_{l,h}\left(\mathbf{h}_{\mathrm{pos}}^{(l,h)} - \mathbf{h}_{\mathrm{neg}}^{(l,h)}\right),
\]
which is exactly the operator $\Phi_s$ with the head-wise strength $s = s^{*}_{l,h}$ of Eq.~\ref{eq:optimal-s}.
In practice HeadTuned-CFG applies one shared $s$ to all selected heads (Section~\ref{sec:method}).

\textbf{Extrapolation and interpolation.}
The vanilla positive branch is $\Phi_s$ with $s=1$, i.e., $q^{*}_{l,h,1} = p_{\mathrm{pos}}^{(l,h)}$, whose expected positive-awareness score is
\[
\mathbb{E}_{p_{\mathrm{pos}}^{(l,h)}}\!\left[T_{l,h}(\mathbf{x})\right]
= \mathbf{w}_{l,h}^{\top}\boldsymbol{\mu}_{\mathrm{neg}}^{(l,h)} + \mathbf{w}_{l,h}^{\top}\Sigma_{l,h}\mathbf{w}_{l,h}.
\]
By Eq.~\ref{eq:optimal-s}, extrapolation ($s^{*}_{l,h} > 1$) is equivalent to
$\tau_{l,h} > \mathbb{E}_{p_{\mathrm{pos}}^{(l,h)}}[T_{l,h}(\mathbf{x})]$, and interpolation ($s^{*}_{l,h} < 1$) to
$\tau_{l,h} < \mathbb{E}_{p_{\mathrm{pos}}^{(l,h)}}[T_{l,h}(\mathbf{x})]$: the target lies above or below the positive-awareness level of the
unchanged positive branch.

HeadTuned-CFG ranks heads by the condition-change awareness score $\delta^{(l,h)} = \lVert \mathbf{h}_{\mathrm{pos}}^{(l,h)} - \mathbf{h}_{\mathrm{neg}}^{(l,h)} \rVert_2$
to separate condition-change aware heads from condition-change agnostic heads, and the ranking guides the assignment of the head-wise targets
$\tau_{l,h}$.
Heads with large $\delta^{(l,h)}$ are condition-change aware; assigning them targets above their unchanged positive-branch level gives
$s^{*}_{l,h} > 1$, the top-$k$ extrapolation mode, which amplifies condition-specific information.
Heads with small $\delta^{(l,h)}$ are condition-change agnostic; assigning them lower targets gives $s^{*}_{l,h} < 1$, the bottom-$k$
interpolation mode, which suppresses their uninformative response and keeps them close to the negative branch.

\section{Implementation Details of HeadTuned-CFG}
\label{app:headtuning-impl}

\textbf{Tuning operator.} For a head $(l,h) \in \mathcal{S}_k$, $\Phi_s$ acts on the per-head slice of an activation tensor and is the linear map
\begin{equation}
    \Phi_s\!\left(\mathbf{h}_{\mathrm{pos}}^{(l,h)}, \mathbf{h}_{\mathrm{neg}}^{(l,h)}\right)
    = \mathbf{h}_{\mathrm{neg}}^{(l,h)} + s\,\bigl(\mathbf{h}_{\mathrm{pos}}^{(l,h)} - \mathbf{h}_{\mathrm{neg}}^{(l,h)}\bigr).
    \label{eq:phi}
\end{equation}
The two branches are run in one batch, so $\mathbf{h}_{\mathrm{neg}}^{(l,h)}$ is available at the same layer.
$\Phi_s$ is implemented as a forward hook that replaces $\mathbf{h}_{\mathrm{pos}}^{(l,h)}$ in place; the negative branch and all downstream
computation, including the CFG combination in Eq.~\ref{eq:cfg}, are unchanged.
$\Phi_s$ acts on the activation probed in Appendix~\ref{app:probing-impl}, $\mathbf{V}_{l,h}$ for Muddit and $\mathbf{O}_{l,h}$ for Janus; the tuned layers,
the token positions it covers, and the tuning steps $\mathcal{T}'$ are chosen per architecture as follows.

\textbf{Muddit.} The 28 single-stream blocks are tuned, restricted to the generation positions; the condition positions and the dual-stream
blocks are left untouched.
Since every denoising step recomputes the full sequence, the tuning steps are $\mathcal{T}' = \mathcal{T}$, and $\mathcal{S}_k$ is re-selected at
each step from the per-step scores.

\textbf{Janus.} All 24 layers are tuned.
The tuning steps are $\mathcal{T}' = \{\text{prefill}\}$: $\Phi_s$ is applied only in the prefill pass and hence only at the condition positions.
The tuned outputs change the hidden states of the condition positions, whose keys and values in the following layers are stored in the KV cache and
attended to by all decode steps, which run unmodified.

\section{Full Metrics on MJHQ-full}
\label{app:full-metrics}

This appendix reports the scores omitted from Tables~\ref{tab:t2i-muddit} and \ref{tab:t2i-janus} on MJHQ-full ($n=30{,}000$), computed
against all 30{,}000 real images: CLIP score~\citep{radford2021learningtransferablevisualmodels} (text--image alignment),
ImageReward~\citep{xu2023imagerewardlearningevaluatinghuman} and HPSv2~\citep{wu2023humanpreferencescorev2} (learned human preference),
precision/recall~\citep{kynkaanniemi2019improvedprecisionrecallmetric} and density/coverage~\citep{naeem2020reliablefidelitydiversitymetrics}
(fidelity and diversity of the generated distribution), Inception Score~\citep{salimans2016improvedtechniquestraininggans}, and the aesthetic
score of the LAION aesthetic predictor~\citep{schuhmann2022laion5bopenlargescaledataset}.

\subsection{Muddit}
\label{app:full-metrics-muddit}

Table~\ref{tab:muddit-overall} lists the scores of vanilla CFG, PAG, PAG+CFG, and HeadTuned-CFG on Muddit; HeadTuned-CFG is the best on
every score except the aesthetic score, where it is within $0.07$ of the best.

\begin{table}[!htb]
\centering
\footnotesize
\setlength{\tabcolsep}{4pt}
\caption{Muddit on MJHQ-full, overall ($n=30{,}000$): CFG ($w=3$), PAG ($\lambda=2.5$ on single-stream blocks 9--10), PAG+CFG
($\lambda=1.5$ on top of CFG with $w=3$), and HeadTuned-CFG ($w=9$, $k=50\%$, $s=0.4$). IR is ImageReward; HPS is human preference score;
Aes.\ is aesthetic score; P/R is precision/recall; D/C is density/coverage.}
\label{tab:muddit-overall}
\begin{tabular}{l cccccccc}
\toprule
& FID $\downarrow$ & CLIP $\uparrow$ & IR $\uparrow$ & HPS $\uparrow$ & P/R $\uparrow$ & D/C $\uparrow$ & IS $\uparrow$ & Aes. $\uparrow$ \\
\midrule
CFG & 20.90 & 0.320 & 0.63 & 0.257 & 0.733/0.464 & 0.856/0.733 & 21.83 & 5.85 \\
PAG & 34.37 & 0.294 & $-0.26$ & 0.217 & 0.597/0.382 & 0.577/0.508 & 16.45 & 5.66 \\
PAG+CFG & 22.39 & 0.318 & 0.64 & 0.264 & 0.720/0.431 & 0.851/0.712 & 20.77 & \textbf{5.91} \\
HeadTuned-CFG & \textbf{16.29} & \textbf{0.324} & \textbf{0.82} & \textbf{0.271} & \textbf{0.762}/\textbf{0.529} & \textbf{0.969}/\textbf{0.808} & \textbf{22.77} & 5.84 \\
\bottomrule
\end{tabular}
\end{table}

\subsection{Janus}
\label{app:full-metrics-janus}
\label{app:janus-pareto}

\begin{figure}[H]
\centering
\includegraphics[width=0.68\linewidth]{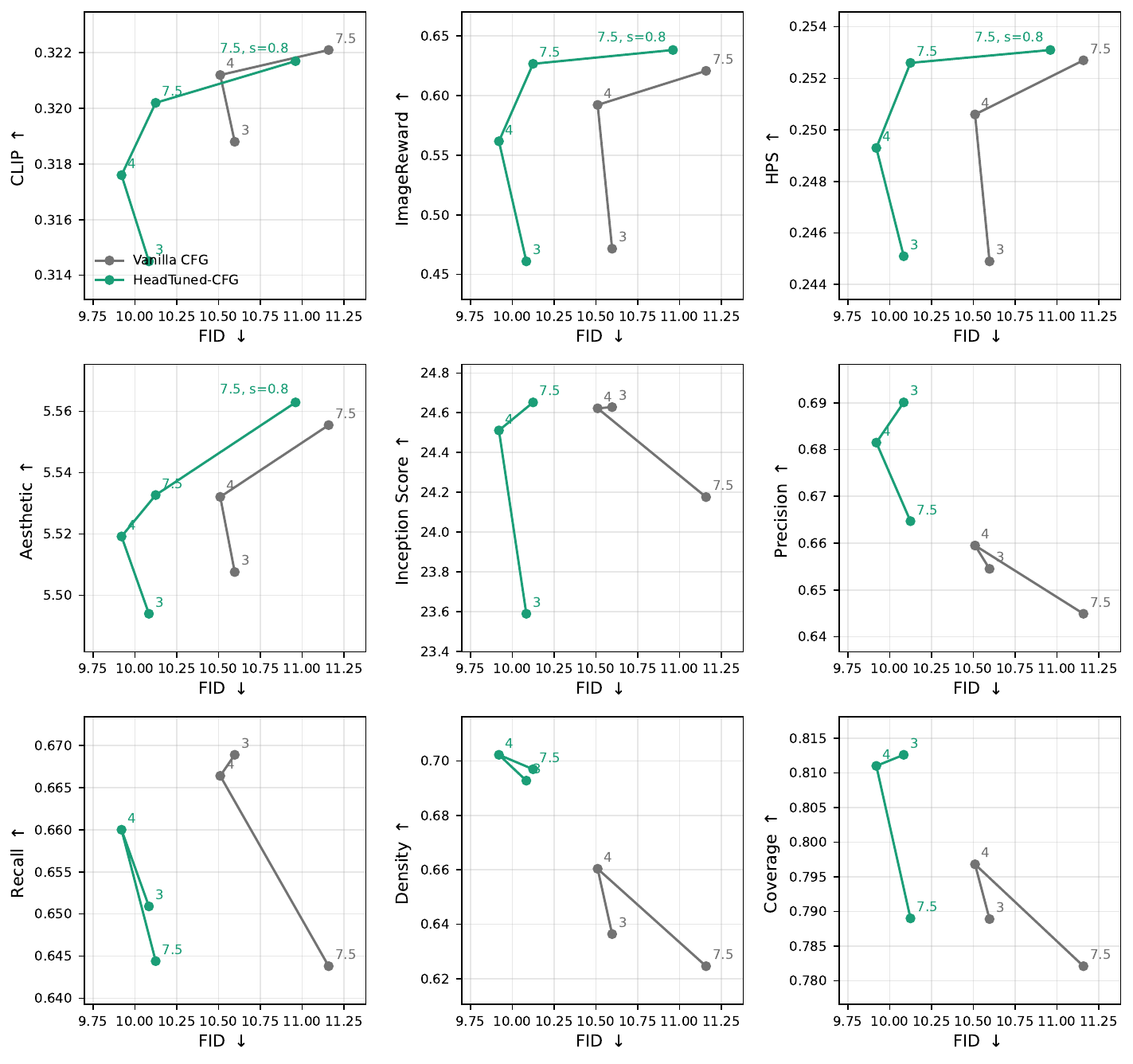}
\caption{Janus on MJHQ-full ($n=30{,}000$): FID against CLIP, ImageReward, HPS, aesthetic score, Inception Score, precision, recall, density,
and coverage over the guidance scale $w$ (labels), for vanilla CFG and HeadTuned-CFG ($k=15\%$, $s=0.75$ unless noted).}
\label{fig:janus-pareto}
\end{figure}

Table~\ref{tab:janus-overall} lists the scores of vanilla CFG and HeadTuned-CFG on Janus: HeadTuned-CFG lowers the FID while the other
scores remain comparable.
To show that the lower FID is not obtained at the expense of the other scores, Figure~\ref{fig:janus-pareto} sweeps $w \in \{3, 4, 7.5\}$ on MJHQ-full for vanilla CFG and for HeadTuned-CFG with the pattern of
Table~\ref{tab:t2i-janus} ($k=15\%$, $s=0.75$) fixed; the CLIP, ImageReward, HPS, and aesthetic panels also include HeadTuned-CFG at $w=7.5$, $s=0.8$.
In every panel, the HeadTuned-CFG curve is Pareto-optimal over the vanilla CFG curve.

\begin{table}[!htb]
\centering
\footnotesize
\setlength{\tabcolsep}{4pt}
\caption{Janus on MJHQ-full, overall ($n=30{,}000$): CFG ($w=4$) versus HeadTuned-CFG ($w=4$, $k=15\%$, $s=0.75$); abbreviations as in
Table~\ref{tab:muddit-overall}.}
\label{tab:janus-overall}
\begin{tabular}{l cccccccc}
\toprule
& FID $\downarrow$ & CLIP $\uparrow$ & IR $\uparrow$ & HPS $\uparrow$ & P/R $\uparrow$ & D/C $\uparrow$ & IS $\uparrow$ & Aes. $\uparrow$ \\
\midrule
CFG & 10.51 & \textbf{0.321} & \textbf{0.59} & \textbf{0.251} & 0.660/\textbf{0.666} & 0.660/0.797 & \textbf{24.62} & \textbf{5.53} \\
HeadTuned-CFG & \textbf{9.92} & 0.318 & 0.56 & 0.249 & \textbf{0.682}/0.660 & \textbf{0.702}/\textbf{0.811} & 24.51 & 5.52 \\
\bottomrule
\end{tabular}
\end{table}

\section{Ablation on Head Selection}
\label{app:headsel}

\begin{figure}[!htb]
\centering
\includegraphics[width=0.85\linewidth]{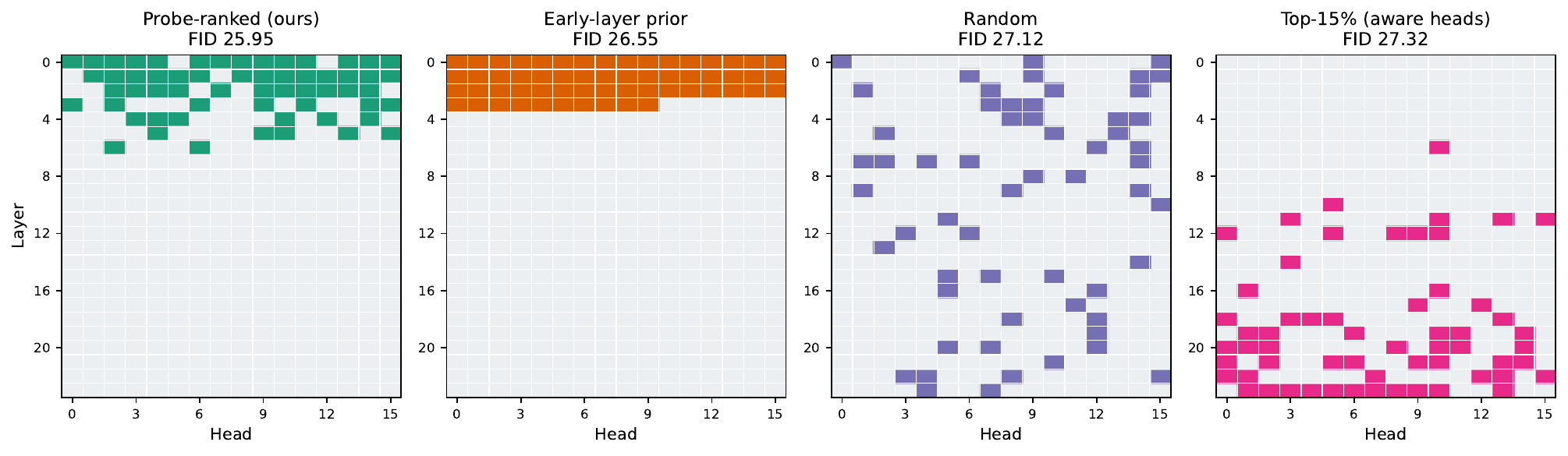}
\caption{Heads tuned by the four selection rules on Janus (layer by head; colored = tuned), with FID on MJHQ-animals at $w=4$,
$k=15\%$, $s=0.75$.}
\label{fig:headsel}
\end{figure}

Figure~\ref{fig:headsel-muddit} in Section~\ref{sec:ablation} shows the head sets of the Muddit ablation; here we report the same
ablation on Janus.
On Janus we keep $w=4$, $k=15\%$, $s=0.75$ and replace the probe-based selection by an early-layer prior (the first layers, where most of
the probed bottom-$15\%$ heads lie), a random set of the same size, and the reverse selection that tunes the top-$15\%$ heads
(Figure~\ref{fig:headsel}).
As on Muddit, the FID ranks probe-based ($25.95$) $<$ early-layer prior ($26.55$) $<$ random ($27.12$) $<$ reverse ($27.32$), against
$27.32$ for vanilla CFG: the reverse selection is no better than no tuning.

\section{Grid Search of the Best Head Pattern under Fixed $w$}
\label{app:grid}
\begingroup
\setlength{\intextsep}{5pt}

\subsection{T2I Generation}
\label{app:grid-t2i}

Resolution and generation steps follow Appendix~\ref{app:probing-t2i}; per-category scores are computed against the real MJHQ images of the
same category.

\subsubsection{Muddit}
\label{app:grid-muddit}

Table~\ref{tab:muddit-vanilla-w} sweeps vanilla CFG for T2I generation on MJHQ-animals; Tables~\ref{tab:muddit-grid-w9} and
\ref{tab:muddit-grid-w3} give the grid search of $(k, s)$ at $w=9$ and $w=3$.

\begin{table}[H]
\centering
\footnotesize
\caption{Vanilla CFG on Muddit, MJHQ-animals ($n=3{,}000$), over the guidance scale $w$.}
\label{tab:muddit-vanilla-w}
\begin{tabular}{l cccccccc}
\toprule
$w$ & 1.0 & 1.5 & 2.0 & 2.5 & 3.0 & 5.0 & 7.0 & 9.0 \\
\midrule
FID & 77.96 & 41.95 & 36.17 & 34.51 & \textbf{33.91} & 34.05 & 35.35 & 37.40 \\
\bottomrule
\end{tabular}
\end{table}

\begin{table}[H]
\centering
\footnotesize
\caption{HeadTuned-CFG search on Muddit, MJHQ-animals, $w=9$: FID over the head fraction $k$ (rows) and strength $s$ (columns).
Vanilla CFG at $w=9$: FID 37.40.}
\label{tab:muddit-grid-w9}
\begin{tabular}{l ccccc}
\toprule
$k \backslash s$ & 0.2 & 0.3 & 0.4 & 0.5 & 0.6 \\
\midrule
40\% & 31.60 & 31.33 & 31.91 & -- & -- \\
45\% & -- & 31.11 & 30.83 & 31.51 & -- \\
50\% & -- & 30.61 & \textbf{30.41} & 31.21 & -- \\
55\% & -- & -- & 31.63 & 31.01 & 31.86 \\
60\% & -- & -- & 31.77 & 31.76 & 32.25 \\
\bottomrule
\end{tabular}
\end{table}

\begin{table}[H]
\centering
\footnotesize
\caption{HeadTuned-CFG search on Muddit, MJHQ-animals, $w=3$: FID over the head fraction $k$ (rows) and strength $s$ (columns). Vanilla CFG at $w=3$: FID 33.91.}
\label{tab:muddit-grid-w3}
\begin{tabular}{l cccc}
\toprule
$k \backslash s$ & 0.6 & 0.7 & 0.8 & 0.9 \\
\midrule
15\% & -- & -- & 33.87 & \textbf{33.61} \\
25\% & -- & 34.09 & 33.70 & 33.80 \\
35\% & 34.57 & 33.83 & 33.98 & 34.00 \\
\bottomrule
\end{tabular}
\end{table}

\subsubsection{Janus}
\label{app:grid-janus}

Table~\ref{tab:janus-vanilla-w} sweeps vanilla CFG for T2I generation on MJHQ-animals; Tables~\ref{tab:janus-grid-w4} and
\ref{tab:janus-grid-w75} give the grid search of $(k, s)$ at $w=4$ and $w=7.5$.

\begin{table}[H]
\centering
\footnotesize
\caption{Vanilla CFG on Janus, MJHQ-animals ($n=3{,}000$), over the guidance scale $w$.}
\label{tab:janus-vanilla-w}
\begin{tabular}{l cccccc}
\toprule
$w$ & 3.0 & 3.5 & 4.0 & 4.5 & 5.0 & 7.5 \\
\midrule
FID & 27.45 & 28.07 & \textbf{27.32} & 27.57 & 27.92 & 28.87 \\
\bottomrule
\end{tabular}
\end{table}

\begin{table}[H]
\centering
\footnotesize
\caption{HeadTuned-CFG search on Janus, MJHQ-animals, $w=4$: FID over the head fraction $k$ (rows) and strength $s$ (columns).
Vanilla CFG at $w=4$: FID 27.32.}
\label{tab:janus-grid-w4}
\begin{tabular}{l ccccc}
\toprule
$k \backslash s$ & 0.60 & 0.65 & 0.70 & 0.75 & 0.80 \\
\midrule
10\% & 26.51 & 26.57 & 26.34 & 27.09 & -- \\
15\% & -- & 26.65 & 26.04 & \textbf{25.95} & 26.92 \\
20\% & -- & 28.12 & 27.08 & 26.32 & 27.19 \\
25\% & -- & 29.74 & 27.23 & 26.40 & 27.22 \\
30\% & -- & -- & 28.88 & 26.95 & 27.35 \\
\bottomrule
\end{tabular}
\end{table}

\begin{table}[H]
\centering
\footnotesize
\caption{HeadTuned-CFG search on Janus, MJHQ-animals, $w=7.5$: FID over the head fraction $k$ (rows) and strength $s$ (columns).
Vanilla CFG at $w=7.5$: FID 28.87.}
\label{tab:janus-grid-w75}
\begin{tabular}{l ccc}
\toprule
$k \backslash s$ & 0.70 & 0.75 & 0.80 \\
\midrule
10\% & 27.11 & 26.76 & 28.01 \\
15\% & 26.95 & \textbf{26.48} & 27.97 \\
20\% & 27.13 & 26.53 & 28.21 \\
\bottomrule
\end{tabular}
\end{table}

\subsection{I2T Generation}
\label{app:grid-i2t}

\textbf{Protocol.} Following VCD, the noised copy of the image is obtained with $T=500$ forward-diffusion steps, and decoding uses the adaptive
plausibility constraint with $\beta=0.2$; each MME subtask has 30 images with two yes/no questions per image and is scored by accuracy plus
image-level accuracy (both questions of an image correct), i.e., out of 200.
Following the official implementation, each POPE question is appended with ``Please answer this question with one word.'' (MME questions
already end with a yes/no instruction), and answers are sampled with temperature 1 and a per-question seed shared by all decoding variants.
Images are encoded at $384 \times 384$ resolution (576 image tokens), as in the I2T probe.
For HeadTuned-VCD, $k$ is given as a number of heads.
The guidance scale of VCD is swept on each benchmark and fixed at its best value, $w=8$ on POPE and $w=10$ on MME; on POPE the head pattern
$(k, s)$ is searched on top of VCD at $w=8$ on the random split, and on MME $(w, k, s)$ is searched jointly over $w \in \{6, 8, 10\}$.

\textbf{POPE.} Table~\ref{tab:janus-vcd-w} sweeps the guidance scale $w$ of VCD on the POPE random split for Janus; VCD saturates from $w=8$ on,
which we take as the baseline. Table~\ref{tab:janus-pope-grid} gives the search of the head pattern on top of VCD at $w=8$.

\begin{table}[H]
\centering
\footnotesize
\caption{VCD on Janus, POPE random split ($n=3{,}000$): accuracy and F1 (\%) over the guidance scale $w=1+\alpha$ ($\beta=0.2$, $T=500$).
Regular decoding: accuracy 84.17, F1 81.60.}
\label{tab:janus-vcd-w}
\begin{tabular}{l cccccc}
\toprule
$w$ & 2 & 4 & 6 & 8 & 10 & 12 \\
\midrule
Acc & 87.10 & 87.60 & 87.70 & \textbf{87.77} & \textbf{87.77} & \textbf{87.77} \\
F1 & 85.47 & 86.12 & 86.25 & \textbf{86.33} & \textbf{86.33} & \textbf{86.33} \\
\bottomrule
\end{tabular}
\end{table}

\begin{table}[H]
\centering
\footnotesize
\caption{HeadTuned-VCD search on Janus, POPE random split, on top of VCD at $w=8$ (accuracy 87.77, F1 86.33): accuracy and F1 (\%) over
the number of tuned heads $k$ (rows) and strength $s$ (columns).}
\label{tab:janus-pope-grid}
\begin{tabular}{l cccc @{\hspace{0.8cm}} cccc}
\toprule
& \multicolumn{4}{c}{Acc.} & \multicolumn{4}{c}{F1} \\
\cmidrule(lr){2-5} \cmidrule(lr){6-9}
$k \backslash s$ & 1.2 & 1.3 & 1.4 & 1.5 & 1.2 & 1.3 & 1.4 & 1.5 \\
\midrule
15 & 88.47 & 88.70 & 88.60 & 88.73 & 87.28 & 87.63 & 87.57 & 87.76 \\
20 & 88.63 & \textbf{88.80} & 88.73 & 88.70 & 87.50 & \textbf{87.76} & 87.74 & \textbf{87.76} \\
25 & 88.53 & 88.53 & 88.27 & 88.00 & 87.40 & 87.46 & 87.22 & 87.01 \\
\bottomrule
\end{tabular}
\end{table}

\textbf{MME.} Table~\ref{tab:janus-mme-w} sweeps $w$ of VCD on the hallucination subset of MME: the score jumps from $w=2$ to $w=4$ and
stays on a plateau from $w=4$ on; the best score is reached at $w=10$, with identical predictions at $w=12$, so we take $w=10$ as the baseline.
Tables~\ref{tab:janus-mme-grid-w6}, \ref{tab:janus-mme-grid-w8}, and \ref{tab:janus-mme-grid-w10} give the search of the head pattern on top
of VCD at $w=6$, $8$, and $10$. The score decreases from $s=1.3$ on for every $w$ and $k$; the best point is $w=8$, $k=15$, $s=1.1$, where
VCD itself is not yet saturated. Adjacent points differ by two to four questions, which is the resolution of the benchmark.

\begin{table}[H]
\centering
\footnotesize
\caption{VCD on Janus, hallucination subset of MME: score per subtask and in total over the guidance scale $w=1+\alpha$ ($\beta=0.2$, $T=500$).
Regular decoding: 138.33 / 56.67 / 90.00 / 93.33, total 378.33.}
\label{tab:janus-mme-w}
\begin{tabular}{l cccccc}
\toprule
$w$ & 2 & 4 & 6 & 8 & 10 & 12 \\
\midrule
existence & 145.00 & 160.00 & 165.00 & 165.00 & 170.00 & 170.00 \\
count & 58.33 & 76.67 & 76.67 & 76.67 & 81.67 & 81.67 \\
position & 80.00 & 93.33 & 93.33 & 91.67 & 91.67 & 91.67 \\
color & 115.00 & 130.00 & 130.00 & 123.33 & 123.33 & 123.33 \\
\midrule
total & 398.33 & 460.00 & 465.00 & 456.67 & \textbf{466.67} & \textbf{466.67} \\
\bottomrule
\end{tabular}
\end{table}

\begin{table}[H]
\centering
\footnotesize
\caption{HeadTuned-VCD search on Janus, hallucination subset of MME, on top of VCD at $w=6$: total score over the number of tuned heads $k$
(rows) and strength $s$ (columns). VCD at $w=6$: 465.00.}
\label{tab:janus-mme-grid-w6}
\begin{tabular}{l cccccc}
\toprule
$k \backslash s$ & 1.05 & 1.10 & 1.15 & 1.20 & 1.25 & 1.30 \\
\midrule
10 & 461.67 & 466.67 & 458.33 & 460.00 & 468.33 & 445.00 \\
15 & 465.00 & 471.67 & 460.00 & 470.00 & 465.00 & 458.33 \\
20 & 468.33 & 463.33 & 466.67 & 468.33 & 461.67 & 455.00 \\
\bottomrule
\end{tabular}
\end{table}

\begin{table}[H]
\centering
\footnotesize
\caption{HeadTuned-VCD search on Janus, hallucination subset of MME, on top of VCD at $w=8$: total score over the number of tuned heads $k$
(rows) and strength $s$ (columns). VCD at $w=8$: 456.67.}
\label{tab:janus-mme-grid-w8}
\begin{tabular}{l cccccc}
\toprule
$k \backslash s$ & 1.05 & 1.10 & 1.15 & 1.20 & 1.25 & 1.30 \\
\midrule
10 & 465.00 & 458.33 & 461.67 & 460.00 & 461.67 & 443.33 \\
15 & 473.33 & \textbf{475.00} & 460.00 & 465.00 & 463.33 & 458.33 \\
20 & 466.67 & 466.67 & 463.33 & 468.33 & 460.00 & 450.00 \\
\bottomrule
\end{tabular}
\end{table}

\begin{table}[H]
\centering
\footnotesize
\caption{HeadTuned-VCD search on Janus, hallucination subset of MME, on top of VCD at $w=10$: total score over the number of tuned heads $k$
(rows) and strength $s$ (columns). VCD at $w=10$: 466.67.}
\label{tab:janus-mme-grid-w10}
\begin{tabular}{l cccccc}
\toprule
$k \backslash s$ & 1.05 & 1.10 & 1.15 & 1.20 & 1.25 & 1.30 \\
\midrule
10 & 470.00 & 463.33 & 461.67 & 460.00 & 461.67 & 443.33 \\
15 & 468.33 & 470.00 & 465.00 & 465.00 & 458.33 & 448.33 \\
20 & 466.67 & 466.67 & 458.33 & 463.33 & 460.00 & 445.00 \\
\bottomrule
\end{tabular}
\end{table}

\endgroup

\end{document}